\documentclass[11pt,letterpaper]{article}
\usepackage{naaclhlt2016}
\usepackage{times}
\naaclfinalcopy

\newif\ifdraft
\drafttrue

\usepackage{amssymb}
\usepackage{amsmath}
\usepackage{array}
\usepackage{color}
\usepackage{dcolumn}
\usepackage{graphicx}
\usepackage{latexsym}
\usepackage{multicol}
\usepackage{multirow}
\usepackage{rotating}
\usepackage{url}
\usepackage[usenames]{xcolor}
\usepackage[utf8]{inputenc}

\title{SemEval-2016 Task 4: Sentiment Analysis in Twitter}

\author{Preslav Nakov$^{\clubsuit}$, Alan Ritter$^{\diamondsuit}$, Sara Rosenthal$^{\heartsuit}$, Fabrizio Sebastiani$^{\clubsuit}$\thanks{\hspace{0.5em}Fabrizio Sebastiani is currently on leave from Consiglio Nazionale delle Ricerche, Italy.}, Veselin Stoyanov$^{\spadesuit}$ \\
$^{\clubsuit}$Qatar Computing Research Institute,
Hamad bin Khalifa University, Qatar \\
$^{\diamondsuit}$Department of Computer Science and Engineering, The Ohio State University, USA \\
$^{\heartsuit}$IBM Watson Health Research, USA \\
$^{\spadesuit}$Johns Hopkins University, USA}

\date{\today}

\begin{document}
\maketitle


\begin{abstract}
  This paper discusses the fourth year of the "Sentiment Analysis in Twitter Task". 
  SemEval-2016 Task 4 comprises five
  subtasks, three of which represent a significant departure from
  previous editions. The first two subtasks are reruns from prior years and ask to predict the overall sentiment, and the sentiment towards a topic in a tweet. The three new subtasks focus on two variants of the basic ``sentiment classification
  in Twitter'' task. 
  The first variant adopts a five-point scale, which confers
  an \textit{ordinal} character to the classification task.  The
  second variant focuses on the correct estimation of the prevalence
  of each class of interest, a task which has been called
  \textit{quantification} in the supervised learning literature. 
  The task continues to be very popular, attracting a total of 43 teams.
\end{abstract}





\section{Introduction}
\label{sec:intro}

\noindent Sentiment classification is the task of detecting whether a
textual item (e.g., a product review, a blog post, an editorial, etc.)
expresses a \textsc{Positive} or a \textsc{Negative} opinion in general or about a given entity, e.g.,
a product, a person, a political party, or a policy. 
Sentiment classification has become
a ubiquitous enabling technology in the Twittersphere. Classifying
tweets according to sentiment has many applications in political
science, social sciences, market research, and many others
\cite{Martinez-Camara:2014qd,Mejova:2015th}. 

As a testament to the prominence of research on sentiment analysis
in Twitter, the tweet sentiment classification (TSC)
task has attracted the highest number of participants in the last three SemEval 
campaigns
\cite{Nakov:2013ty,Rosenthal:2014ty,Rosenthal:2015ty,Nakov:2016rm}.

Previous editions of the SemEval task involved binary
(\textsc{Positive} vs.\ \textsc{Negative}) or \emph{single-label multi-class} classification (SLMC) when a \textsc{Neutral}\footnote{
We merged \textsc{Objective} under \textsc{Neutral}, as previous attempts to have
annotators distinguish between the two have consistently resulted in
very low inter-annotator agreement.} 
class is added (\textsc{Positive}
vs.\ \textsc{Negative} vs.\ \textsc{Neutral}).
SemEval-2016 Task 4 represents a significant departure from these
previous editions.  Although two of the subtasks (Subtasks A and B)
are reincarnations of previous editions (SLMC classification for
Subtask A, binary classification for Subtask B), SemEval-2016 Task 4
introduces two completely new problems, taken individually (Subtasks C
and D) and in combination (Subtask E):

\subsection{Ordinal Classification}

We replace the 
two- or three-point scale with a
  five-point scale \{\textsc{HighlyPositive}, \textsc{Positive},
  \textsc{Neutral}, \textsc{Negative}, \textsc{HighlyNegative}\}, which is now
  ubiquitous in the corporate world where human ratings are involved:
  e.g., Amazon, TripAdvisor, and Yelp, 
  all use a five-point scale for rating sentiment towards products, hotels, and
  restaurants.

  Moving from a categorical two/three-point scale to an ordered
  five-point scale means, in machine learning terms, moving from
  binary to \textit{ordinal classification} (a.k.a.\
  \textit{ordinal regression}).

\subsection{Quantification} 

We replace classification with \textit{quantification}, i.e.,
  supervised class prevalence estimation. With regard to Twitter, hardly anyone is
  interested in whether \textit{a specific person} has a positive or a
  negative view of the topic. Rather, applications look at  estimating the \textit{prevalence} of positive
  and negative tweets about a given topic. Most (if not all) tweet sentiment classification studies conducted within political science
  \cite{Borge-Holthoefer:2015dz,Kaya:2013ca,marchettibowick-chambers:2012:EACL2012},
  economics \cite{Bollen:2011bf,OConnor:2010fk}, social science
  \cite{Dodds:2011uq}, and market research
  \cite{Burton:2011sh,Qureshi:2013fb}, use Twitter with an interest in
  aggregate data and \emph{not} in individual classifications.

  Estimating prevalences (more generally, estimating the
  \textit{distribution} of the classes in a set of unlabelled items)
  by leveraging training data is called \textit{quantification} in
  data mining and related fields. Previous work has argued that
  quantification is not a mere byproduct of classification, since (a)
  a good classifier is not necessarily a good quantifier, and vice
  versa, see, e.g., \cite{Forman:2008kx}; (b) quantification requires
  evaluation measures different from classification. Quantification-specific learning approaches have been
  proposed over the years; Sections 2 and 5 of \cite{Esuli:2015gh}
  contain several pointers to such literature.

Note that, in Subtasks B to E, tweets come labelled with the
\textit{topic} they are about and participants need not classify whether a tweet is about a given topic. A
topic can be anything that people express opinions
about; for example, a product (e.g., iPhone6), a political candidate
(e.g., Hillary Clinton), a policy (e.g., Obamacare), an event (e.g.,
the Pope's visit to Palestine), etc.

The rest of the paper is structured as follows. In Section
\ref{sec:task}, we give a general overview of SemEval-2016 Task 4 and 
the five subtasks.
Section~\ref{sec:datasets} focuses on the datasets,
and on the data generation procedure. In Section \ref{sec:scoring},
we describe in detail the evaluation measures for each subtask.
Section \ref{sec:results} discusses the results of
the evaluation and the techniques and tools that the
top-ranked participants used. Section \ref{sec:conclusion}
concludes, discussing the lessons learned and some possible ideas for a followup at SemEval-2017.


\section{Task Definition}
\label{sec:task}

\noindent SemEval-2016 Task 4 consists of five subtasks:

\begin{enumerate}
\item \textbf{Subtask A:} Given a tweet, predict whether it is of
  positive, negative, or neutral sentiment.
\item \textbf{Subtask B:} Given a tweet known to be about a given topic,
  predict whether it conveys a positive or a negative sentiment
  towards the topic.
\item \textbf{Subtask C:} Given a tweet known to be about a given topic,
  estimate the sentiment it conveys towards the topic on a
  five-point scale ranging from \textsc{HighlyNegative} to
  \textsc{HighlyPositive}.
\item \textbf{Subtask D:} Given a set of tweets known to be about a given
  topic, estimate the distribution of the tweets in the
  \textsc{Positive} and \textsc{Negative} classes.

\item \textbf{Subtask E:} Given a set of tweets known to be about a given
  topic, estimate the distribution of the tweets across the five
  classes of a five-point scale, ranging from \textsc{HighlyNegative}
  to \textsc{HighlyPositive}.

\end{enumerate}

\noindent Subtask A is a rerun -- it was present in all three previous
editions of the task. In the 2013-2015 editions, it was known as
Subtask B.\footnote{Note that we retired the expression-level subtask A, which was present in SemEval 2013--2015 \cite{Nakov:2013ty,Rosenthal:2014ty,Rosenthal:2015ty,Nakov:2016rm}.} We ran it again this year because it was the most popular subtask in
the three previous task editions. It was the most popular subtask this
year as well -- see Section \ref{sec:results}.

Subtask B is a variant of SemEval-2015 Task 10 Subtask C \cite{Rosenthal:2015ty,Nakov:2016rm}, with
\textsc{Positive}, \textsc{Neutral}, and \textsc{Negative} as the
classification labels.

Subtask E is similar to SemEval-2015 Task 10 Subtask D, which
consisted of the following problem: \textit{Given a set of messages on
a given topic from the same period of time, classify the overall
sentiment towards the topic in these messages as strongly positive,
weakly positive, neutral, weakly negative, or strongly negative}. Note
that in SemEval-2015 Task 10 Subtask D, exactly one of the five classes
had to be chosen, while in our Subtask E, a distribution across the
five classes has to be estimated.

As per the above discussion, Subtasks B to E are new. 
Conceptually, they form a 2$\times$2 matrix,
as shown in Table~\ref{tab:tasks}, where the rows indicate the \emph{goal}
of the task (classification vs.\ quantification) and the columns
indicate the \emph{granularity} of the task (two- vs.\ five-point scale).

\begin{table}[ht!]
  \begin{center}
    \resizebox{\columnwidth}{!} {
    \begin{tabular}{|c|c||c|c|}
      \hline
      & & \multicolumn{2}{c|}{Granularity} \\
      \cline{3-4}
      & & Two-point & Five-point \\
      & & (binary) & (ordinal) \\
      \hline\hline
      \multirow{2}{*}{\begin{sideways}Goal\end{sideways}} & Classification & Subtask B & Subtask C \\
      \cline{2-4}
      & Quantification & Subtask D & Subtask E \\
      \hline
    \end{tabular}
    }
    \caption{\label{tab:tasks}A 2$\times$2 matrix summarizing the
    similarities and the differences between Subtasks B-E.}
  \end{center}
\end{table}%





\section{Datasets}
\label{sec:datasets}

\noindent In this section, we describe the process of collection and
annotation of the training, development and testing tweets for all
five subtasks. We dub this dataset the \emph{Tweet 2016} dataset in
order to distinguish it from datasets generated in previous editions
of the task.


\subsection{Tweet Collection}

We provided the datasets from the previous
editions\footnote{For Subtask A, we did not allow training on the
testing datasets from 2013--2015, as we used them for progress
testing.} (see Table~\ref{T:CorpusStatsB:old:editions}) of this
task~\cite{Nakov:2013ty,Rosenthal:2014ty,Rosenthal:2015ty,Nakov:2016rm}
for training and development. In addition we created new training and testing
datasets. 

\begin{table}[ht!]
  \small
  \begin{center}
    \resizebox{\columnwidth}{!} {
    \begin{tabular}{|l|r r r |r|}
      \hline
      \multicolumn{1}{|c|}{\begin{sideways}\bf Dataset\end{sideways}} & \multicolumn{1}{c}{\begin{sideways}\textsc{Positive}\end{sideways}} & \multicolumn{1}{c}{\begin{sideways}\textsc{Negative~~}\end{sideways}} & \multicolumn{1}{c|}{\begin{sideways}\textsc{neutral}\end{sideways}} & \multicolumn{1}{c|}{\begin{sideways}\bf Total\end{sideways}}\\
      \hline
      Twitter2013-train & 3,662 & 1,466 &  4,600 & 9,728\\
      Twitter2013-dev & 575 & 340 &  739 & 1,654\\
      Twitter2013-test  & 1,572 & 601 & 1,640 & 3,813\\
      SMS2013-test  & 492 & 394 & 1,207 & 2,093\\
      Twitter2014-test  & 982 & 202 & 669 & 1,853\\
      \footnotesize{Twitter2014-sarcasm} & 33 & 40 & 13 & 86\\
      \footnotesize{LiveJournal2014-test}  & 427 & 304 & 411 & 1,142\\
      Twitter2015-test & 1,040 & 365 & 987 & 2,392 \\
      \hline
    \end{tabular}
    }
    \caption{Statistics about data from the 2013-2015 editions of the SemEval task on Sentiment Analysis in Twitter, which could be used for training and development for SemEval-2016 Task 4.}
    \label{T:CorpusStatsB:old:editions}
  \end{center}
\end{table}

We employed the following annotation procedure.
As in previous years, we first gathered tweets that express sentiment
about popular topics. For this purpose, we extracted named entities
from millions of tweets, using a Twitter-tuned named entity
recognition system \cite{ner}. The collected tweets were greatly
skewed towards the neutral class. In order to reduce the class
imbalance, we removed those that contained no sentiment-bearing
words. We used SentiWordNet 3.0 \cite{Baccianella:2010fk} as a
repository of sentiment words. Any word listed in SentiWordNet 3.0
with at least one sense having a positive or a negative sentiment
score greater than 0.3 was considered sentiment-bearing.\footnote{Filtering based on an existing lexicon does bias the
dataset to some degree; however, the text still contains
sentiment expressions outside those in the lexicon.}

The training and development tweets were collected from July to
October 2015. The test tweets were collected from October to
December 2015. We used the public streaming Twitter API to download
the tweets.\footnote{We distributed the datasets to the task
participants in a similar way: we only released the annotations and
the tweet IDs, and the participants had to download the actual tweets
by themselves via the API, for which we provided a
script: \texttt{https://github.com/aritter/twitter\_download}}

We then manually filtered the resulting tweets to obtain a set of 200
meaningful topics with at least 100 tweets each (after filtering out
near-duplicates).  We excluded topics that were incomprehensible,
ambiguous (e.g., \emph{Barcelona}, which is the name both of a city and of a sports team), or too general (e.g.,~\emph{Paris}, which is the
name of a big city). We then discarded tweets that were just
mentioning the topic but were not really about the topic.

Note that the topics in the training and in the test sets do not overlap, i.e.,
the test set consists of tweets about topics different from the topics
the training and development tweets are about.


\subsection{Annotation}

\begin{figure*}
  \framebox{\parbox[t]{.98\textwidth}{\small\textbf{Instructions:}
  Given a Twitter message and a topic, identify whether the message is
  highly positive, positive, neutral, negative, or highly negative (a)
  in general and (b) with respect to the provided topic. If a tweet is
  sarcastic, please select the checkbox ``The tweet is
  sarcastic". Please read the examples and the invalid responses
  before beginning if this is the first time you are working on this
  HIT.}}
  \framebox{\centering\parbox[t]{.98\textwidth}{\includegraphics[scale=.49]{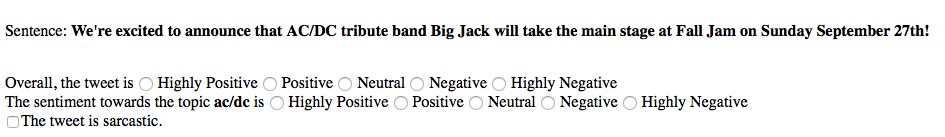}}}
  \caption{The instructions provided to the Mechanical Turk
  annotators, followed by a screenshot.}
  \label{F:instructions}
\end{figure*}

The 2016 data consisted of four parts: TRAIN (for training models),
DEV (for tuning models), DEVTEST (for development-time evaluation),
and TEST (for the official evaluation). The first three datasets were
annotated using Amazon's Mechanical Turk, while the TEST dataset was
annotated on CrowdFlower.

\paragraph{Annotation with Amazon's Mechanical Turk.} A Human
Intelligence Task (HIT) consisted of providing all required
annotations for a given tweet message.
In order to qualify to work on our HITs, a Mechanical Turk annotator
(a.k.a.\ ``Turker'') had to have an approval rate greater than 95\%
and to have completed at least 50 approved HITs. Each HIT was carried
out by five Turkers and consisted of five tweets to be annotated. A Turker had to indicate the overall polarity of the
tweet message (on a five-point scale) as well as the overall
polarity of the message towards the given target topic (again, on a
five-point scale). The annotation instructions along with an
example are shown in Figure~\ref{F:instructions}.  We made
available to the Turkers several additional examples, which are shown
in Table~\ref{T:annotation-examples}.

\begin{table*}[t!]
  \centering
  \resizebox{\textwidth}{!} {
  \begin{tabular}{|p{9cm}|p{1.5cm}|p{5.5cm}|}
    \hline
    \multicolumn{1}{|c|}{\bf Tweet} & \multicolumn{1}{c|}{\bf Overall Sentiment} & \multicolumn{1}{c|}{\bf Topic Sentiment} \\
    \hline\hline
    Why would you still wear shorts when it's this cold?! I love how Britain see's a bit of sun and they're like 'OOOH LET'S STRIP! &	\textsc{Positive} &	Britain: \textsc{Negative} \\
    \hline
    Saturday without Leeds United is like Sunday dinner it doesn't feel normal at all (Ryan)	& \textsc{Negative}	& Leeds United: \textsc{HighlyPositive} \\
    \hline
    Who are you tomorrow? Will you make me smile or just bring me sorrow? \#HottieOfTheWeek Demi Lovato	& \textsc{Neutral}	& Demi Lovato: \textsc{Positive} \\
    \hline
  \end{tabular}
  }
  \caption{List of example tweets and annotations that were provided to the annotators.}
  \label{T:annotation-examples}
\end{table*}

We rejected HITs with the following problems:

\begin{itemize}
\item one or more responses do not have the overall sentiment marked;
\item one or more responses do not have the sentiment towards the
  topic marked;
\item one or more responses appear to be randomly selected.
\end{itemize}

\paragraph{Annotation with CrowdFlower.} 
We annotated the TEST data using CrowdFlower, as it allows better
quality control of the annotations across a number of dimensions. Most
importantly, it allows us to find and exclude unreliable annotators based
on hidden tests, which we created starting with the highest-confidence
and highest-agreement annotations from Mechanical Turk. We added some more tests manually.  Otherwise, we
setup the annotation task giving exactly the same instructions and
examples as in Mechanical Turk.

\paragraph{Consolidation of annotations.} 
In previous years,
we used majority voting to select the true label (and discarded
cases where a majority had not emerged, which amounted to about 50\%
of the tweets). As this year we have a five-point scale, where the
expected agreement is lower, we used a two-step procedure. If three
out of the five annotators agreed on a label, we accepted the label.
Otherwise, we first mapped the categorical labels to the integer
values $-$2, $-$1, 0, 1, 2. Then we calculated the average, and
finally we mapped that average to the closest integer value. In order
to counter-balance the tendency of the average to stay away from
$-$2 and 2, and also to prefer 0, we did not use rounding at $\pm$0.5
and $\pm$1.5, but at $\pm$0.4 and $\pm$1.4 instead.

To give the reader an idea about the degree of agreement, we will look at the TEST dataset as an example. It included 20,632 tweets. For 2,760,
all five annotators assigned the same value, and for another 9,944
there was a majority value. For the remaining 7,928 cases, we had to
perform averaging as described above.

The consolidated statistics from the five annotators on a three-point
scale for Subtask A are shown in Table~\ref{T:CorpusStats}.  Note
that, for consistency, we annotated the data for Subtask A on a
five-point scale, which we then converted to a three-point scale.

\begin{table}[tbh]
  \small
  \begin{center}
    \resizebox{\columnwidth}{!} {
    \begin{tabular}{|l|rrr|r|}
      \hline
      \bf & \multicolumn{1}{c}{\begin{sideways}\textsc{Positive}\end{sideways}} & \multicolumn{1}{c}{\begin{sideways}\textsc{Neutral}\end{sideways}} & \multicolumn{1}{c|}{\begin{sideways}\textsc{Negative~~}\end{sideways}} & \multicolumn{1}{c|}{\begin{sideways}\bf Total\end{sideways}}\\
      \hline
      TRAIN & 3,094 & 863 & 2,043 & 6,000 \\
      DEV & \phantom{0}844 & 765 & \phantom{0}391 & 2,000 \\
      DEVTEST & \phantom{0}994 & 681 & \phantom{0}325 & 2,000 \\
      TEST & 7,059 & 10,342 & 3,231 & 20,632 \\
      \hline
    \end{tabular}
    }
    \caption{2016 data statistics (Subtask A).}
    \label{T:CorpusStats}
  \end{center}
\end{table}

The topic annotations on a two-point scale for Subtasks B and D are
shown in Table~\ref{T:CorpusStatsTopicBD}, while those on a five-point
scale for Subtasks C and E are in
Table~\ref{T:CorpusStatsTopicCE}. Note that, as for Subtask A, the
two-point scale annotation counts for Subtasks B and D derive from
summing the \textsc{Positive}s with the \textsc{HighlyPositive}s, and
the \textsc{Negative}s with the \textsc{HighlyNegative}s from
Table~\ref{T:CorpusStatsTopicCE}; moreover, this time we also remove
the \textsc{Neutral}s.

\begin{table}[th!]
  \small
  \begin{center}
    \resizebox{\columnwidth}{!} {
    \begin{tabular}{|l|r|rr|r|}
      \hline
      & \multicolumn{1}{c|}{\begin{sideways}\bf Topics\end{sideways}} & \multicolumn{1}{c}{\begin{sideways}\textsc{Positive}\end{sideways}} & \multicolumn{1}{c|}{\begin{sideways}\textsc{Negative~~}\end{sideways}} & \multicolumn{1}{c|}{\begin{sideways}\bf Total\end{sideways}}\\
      \hline
      TRAIN & 60 & 3,591 & 755 & 4,346 \\
      DEV & 20 & \phantom{0}986 & 339 & 1,325 \\
      DEVTEST & 20 & 1,153 & 264 & 1,417 \\
      TEST & 100 & 8,212 & 2,339 & 10,551\\
      \hline
    \end{tabular}
    }
    \caption{2016 data statistics (Subtasks B and D).}
    \label{T:CorpusStatsTopicBD}
  \end{center}
\end{table}

\begin{table}[tbh!]
  \small
  \begin{center}
    \resizebox{\columnwidth}{!} {
    \begin{tabular}{|l|r|rrrrr|r|}
      \hline
      & \multicolumn{1}{c|}{\begin{sideways}\bf Topics\end{sideways}} & \multicolumn{1}{c}{\begin{sideways}\textsc{HighlyPositive}\end{sideways}} & \multicolumn{1}{c}{\begin{sideways}\textsc{Positive}\end{sideways}} & \multicolumn{1}{c}{\begin{sideways}\textsc{Neutral}\end{sideways}} & \multicolumn{1}{c}{\begin{sideways}\textsc{Negative}\end{sideways}} & \multicolumn{1}{c|}{\begin{sideways}\textsc{HighlyNegative~~~}\end{sideways}} & \multicolumn{1}{c|}{\begin{sideways}\bf Total\end{sideways}}\\
      \hline
      TRAIN & 60 & 437 & 3,154 & 1,654 & 668 & 87 & 6,000 \\
      DEV & 20 & \phantom{0}53 & \phantom{0}933 & \phantom{0}675 & 296 & 43 & 2,000 \\
      DEVTEST & 20 & 148 & 1,005 & \phantom{0}583 & 233 & 31 & 2,000 \\
      TEST & 100 & 382 & 7,830 & 10,081 & 2,201 & 138 & 20,632 \\
      \hline
    \end{tabular}
    }
    \caption{2016 data statistics (Subtasks C and E).}
    \label{T:CorpusStatsTopicCE}
  \end{center}
\end{table}

As we use the same test tweets for all subtasks, the submission of results
by participating teams was subdivided in two stages: (\emph{i}) participants had to submit results
for Subtasks A, C, E, and (\emph{ii}) only after the submission deadline for A, C, E had
passed, we distributed to participants the unlabelled test data for Subtasks B and D. 

Otherwise, since for Subtasks B and D we filter out
the \textsc{Neutral}s, we would have leaked information about which
the \textsc{Neutral}s are, and this information could have been used
in Subtasks C and E.

Finally, as the same tweets can be selected for different topics, we
ended up with some duplicates; arguably, these are true duplicates for
Subtask A only, as for the other subtasks the topics still
differ. This includes 25 duplicates in TRAIN, 3 in DEV, 2 in DEVTEST,
and 116 in TEST. There is a larger number in TEST, as TEST is about
twice as large as TRAIN, DEV, and DEVTEST combined. This is because we wanted a large
TEST set with 100 topics and 200 tweets per topic on average for
Subtasks C and E.


\section{Evaluation Measures}
\label{sec:scoring}

\noindent This section discuss the evaluation measures for the five subtasks of
our SemEval-2016 Task 4. A document describing
the evaluation measures in detail\footnote{\texttt{http://alt.qcri.org/semeval2016/task4/}} 
\cite{Nakov:2016kl}, and a scoring software implementing all the five
``official'' measures, were made available to the participants via the task website
together with the training data.\footnote{An earlier version of the
scoring script contained a bug, to the effect that for Subtask B it
was computing $F_1^{PN}$, and not $\rho^{PN}$. This was detected only
after the submissions were closed, which means that participants to
Subtask B who used the scoring system (and not their own
implementation of $\rho^{PN}$) for parameter optimization, may have
been penalized in the ranking as a result.}

For Subtasks B to E, the 
datasets are each subdivided into a number of ``topics'', and
the subtask needs to be carried out independently for each topic. As a
result, each of the evaluation measures will be
``macroaveraged'' across the topics, i.e.,~we compute the measure
individually for each topic, and we then average the results across
the topics.


\subsection{Subtask A: Message polarity classification}
\label{sec:4A}

\noindent Subtask A is a {\em single-label multi-class} (SLMC) classification task. Each tweet must be classified as belonging to
exactly one of the following three classes
$\mathcal{C}$=\{\textsc{Positive}, \textsc{Neutral},
\textsc{Negative}\}.

We adopt the same evaluation measure as the 2013-2015 editions of this subtask,
$F_{1}^{PN}$:
\begin{equation}\label{eq:F1PN}F_{1}^{PN}=\frac{F_{1}^{P}+F_{1}^{N}}{2}\end{equation}
\noindent $F_{1}^{P}$
is the $F_{1}$ score for the \textsc{Positive} class:

\begin{equation}
  \begin{aligned}
    \label{eq:f1pos}
    F_{1}^{P}=\dfrac{2\pi^{P}\rho^{P}}{\pi^{P}+\rho^{P}}
  \end{aligned}
\end{equation}

\noindent Here, $\pi^{P}$ and $\rho^{P}$ denote precision and recall
for the \textsc{Positive} class, respectively:
\begin{equation}
  \begin{aligned}
    \label{eq:pipos}
    \pi^{P} =\frac{PP}{PP+PU+PN}
  \end{aligned}
\end{equation}
\begin{equation}
  \begin{aligned}
    \label{eq:rhopos}
    \rho^{P} =\frac{PP}{PP+UP+NP}
  \end{aligned}
\end{equation}

\noindent where $PP$, $UP$, $NP$, $PU$, $PN$ are the cells of the
confusion matrix shown in Table~\ref{tab:threewaycontingencytable}.

\renewcommand{\tabcolsep}{0.1cm} \renewcommand{\arraystretch}{1.30}
\begin{table}[tbh]
  \begin{center}
    \resizebox{\columnwidth}{!} {
    \begin{tabular}{|c|c||c|c|c|}
      \hline
      \multicolumn{2}{|c||}{\mbox{}} & \multicolumn{3}{c|}{\bf Gold Standard} \\
      \cline{3-5}
      \multicolumn{2}{|c||}{\mbox{}} & \textsc{Positive} & \textsc{Neutral} & \textsc{Negative} \\
      \hline\hline
      \multirow{3}{*}{\begin{sideways}\bf Predicted\end{sideways}} & \textsc{Positive} & PP & PU & PN \\
      \cline{2-5}
      & \textsc{Neutral} & UP & UU & UN \\
      \cline{2-5}
      & \textsc{Negative} & NP & NU & NN \\
      \hline
    \end{tabular}
    }
\caption{
\label{tab:threewaycontingencytable} The confusion matrix for Subtask
A. Cell $XY$ stands for ``the number of tweets that the classifier labeled $X$
and the gold standard labells as $Y$''. $P$, $U$, $N$ stand for
\textsc{Positive}, \textsc{Neutral}, \textsc{Negative}, respectively.}
\end{center}
\end{table}%
$F_{1}^{N}$ is defined analogously, and the measure we finally adopt
is $F_{1}^{PN}$ as from Equation \ref{eq:F1PN}.


\subsection{Subtask B: Tweet classification according to a two-point
scale}
\label{sec:4B}

\noindent Subtask B is a \emph{binary classification} task. Each tweet must be classified as either \textsc{Positive} or \textsc{Negative}.


For this subtask we adopt
%
%
\emph{macroaveraged recall}:
\begin{equation}
  \begin{aligned}
    \label{eq:rhoPN}
    \rho^{PN} & = \frac{1}{2}(\rho^{P}+\rho^{N}) \\
    & =
    \frac{1}{2}(\displaystyle\frac{PP}{PP+NP}+\displaystyle\frac{NN}{NN+PN})
  \end{aligned}
\end{equation}

In the above formula, $\rho^{P}$ and $\rho^{N}$ are the positive and the negative class recall, respectively. Note that $U$ terms are entirely missing in Equation
\ref{eq:rhoPN}; this is because we do not have the \textsc{Neutral} class for SemEval-2016 Task 4, subtask A.

$\rho^{PN}$ ranges in $[0,1]$, where a value of 1 is achieved only by the perfect
classifier (i.e., the classifier that correctly classifies all items), a value of 0 is achieved only by the perverse classifier (the classifier that
misclassifies all items), while $0.5$ is both (\emph{i}) the value obtained
by a trivial classifier (i.e.,~the classifier that assigns all tweets
to the same class -- be it \textsc{Positive} or \textsc{Negative}),
and (\emph{ii}) the expected value of a random classifier. The advantage of
$\rho^{PN}$ over ``standard'' accuracy is that it is more robust to
class imbalance.  The accuracy of the
majority-class classifier is the relative frequency (aka
``prevalence'') of the majority class, that may be much higher than
0.5 if the test set is imbalanced.  Standard $F_{1}$ is also sensitive
to class imbalance for the same reason. Another advantage of
$\rho^{PN}$ over $F_{1}$ is that $\rho^{PN}$ is invariant with respect
to switching \textsc{Positive} with \textsc{Negative}, while $F_{1}$
is not.  See \cite{Sebastiani:2015zl} for more details on $\rho^{PN}$.

As we noted before, the training dataset, the development dataset, and the test
dataset are each subdivided into a number of topics, and Subtask B
needs to be carried out independently for each topic. As a result, the
evaluation measures discussed in this section are computed
individually for each topic, and the results are then averaged across
topics to yield the final score.


\subsection{Subtask C: Tweet classification according to a five-point
scale}
\label{sec:4C}

\noindent Subtask C is an \emph{ordinal classification} (OC -- also
known as \emph{ordinal regression}) task, in which each tweet must be
classified into exactly one of the classes in
$\mathcal{C}$=\{\textsc{HighlyPositive}, \textsc{Positive},
\textsc{Neutral}, \textsc{Negative}, \textsc{HighlyNegative}\},
represented in our dataset by numbers in \{$+2$,$+1$,$0$,$-1$,$-2$\}, with a
total order defined on $\mathcal{C}$. The essential difference between
SLMC (see Section \ref{sec:4A} above) and OC is that not all mistakes weigh
equally in the latter. For example, misclassifying a
\textsc{HighlyNegative} example as \textsc{HighlyPositive} is a bigger
mistake than misclassifying it as \textsc{Negative} or \textsc{Neutral}.

As our evaluation measure, we use \emph{macroaveraged mean absolute
error} ($MAE^{M}$):
\begin{equation}
  \label{eq:macroMAE}MAE^{M}(h,Te) =
  \frac{1}{|\mathcal{C}|}\sum_{j=1}^{|\mathcal{C}|}\frac{1}{|Te_{j}|}\sum_{\mathbf{x}_{i}\in
  Te_{j}}|h(\mathbf{x}_{i})-y_{i}| \nonumber
\end{equation}
\noindent where $y_{i}$ denotes the true label of item
$\mathbf{x}_{i}$, $h(\mathbf{x}_{i})$ is its predicted label,
$Te_{j}$ denotes the set of test documents whose true class is
$c_{j}$, $|h(\mathbf{x}_{i})-y_{i}|$ denotes the ``distance'' between
classes $h(\mathbf{x}_{i})$ and $y_{i}$ (e.g., the distance between
\textsc{HighlyPositive} and \textsc{Negative} is 3), and the ``M''
superscript indicates ``macroaveraging''.

The advantage of $MAE^{M}$ over ``standard'' mean absolute error,
which is defined as:
\begin{equation}\label{eq:microMAE}MAE^{\mu}(h,Te)=\frac{1}{
  |Te|}\sum_{\mathbf{x}_{i}\in
  Te}|h(\mathbf{x}_{i})-y_{i}|\end{equation}
\noindent is that it is robust to class imbalance (which is
useful, given the imbalanced nature of our dataset). On perfectly
balanced datasets $MAE^{M}$ and $MAE^{\mu}$ are equivalent.

Unlike the measures discussed in Sections \ref{sec:4A} and
\ref{sec:4B}, $MAE^{M}$ is a measure of error, and not
accuracy, and thus lower values are better. See~\cite{Baccianella:2009qd}
for more detail on $MAE^{M}$.

Similarly to Subtask B, Subtask C needs to be carried out independently for
each topic. As a result, $MAE^{M}$ is computed individually for each
topic, and the results are then averaged across all topics to yield
the final score.


\subsection{Subtask D: Tweet quantification according to a two-point
scale}
\label{sec:4D}

\noindent Subtask D also assumes a \emph{binary quantification} setup,
in which each tweet is classified as
\textsc{Positive} or \textsc{Negative}. The task is to
compute an estimate $\hat{p}(c_{j})$ of
the relative frequency (in the test set) of each of the
classes.

The difference between binary classification (as from Section
\ref{sec:4B}) and binary quantification is that errors of different
polarity (e.g., a false positive and a false negative for the same
class) can compensate each other in the latter. Quantification is thus
a more lenient task
since a perfect classifier is also a perfect quantifier,
but a perfect quantifier is not necessarily a perfect classifier.

We adopt \emph{normalized cross-entropy},
better known as \emph{Kullback-Leibler Divergence} (KLD). KLD was
proposed as a quantification measure in \cite{Forman:2005fk}, and is
defined as follows:
\begin{equation}
  \label{eq:KLD}
  KLD(\hat{p},p,\mathcal{C}) = \sum_{c_{j}\in \mathcal{C}} 
  p(c_{j})\log_e\frac{p(c_{j})}{\hat{p}(c_{j})}
\end{equation}

$KLD$ is a measure of the error made in estimating a true
distribution $p$ over a set $\mathcal{C}$ of classes by means of a
predicted distribution $\hat{p}$. Like $MAE^{M}$ in Section
\ref{sec:4C}, $KLD$ is a measure of error, which means that lower values are
better. $KLD$ ranges between 0 (best) and $+\infty$ (worst).

Note that the upper bound of $KLD$ is not finite since Equation
\ref{eq:KLD} has predicted prevalences, and not true prevalences, at
the denominator: that is, by making a predicted prevalence
$\hat{p}(c_{j})$ infinitely small we can make $KLD$ infinitely large.
To solve this problem, in computing $KLD$ we smooth both $p(c_{j})$
and $\hat{p}(c_{j})$ via additive smoothing, i.e.,
\begin{equation}
  \begin{aligned}
    \label{eq:smoothing}
    p^{s}(c_{j})= &
    \frac{p(c_{j})+\epsilon}{(\displaystyle\sum_{c_{j}\in
    \mathcal{C}}p(c_{j}))+\epsilon\cdot|\mathcal{C}|} \\ = &
    \frac{p(c_{j})+\epsilon}{1+\epsilon\cdot|\mathcal{C}|}
  \end{aligned}
\end{equation}

\noindent where $p^{s}(c_{j})$ denotes the smoothed version of
$p(c_{j})$ and the denominator is just a normalizer (same for
the $\hat{p}^{s}(c_{j})$'s); the quantity $\epsilon=\frac{1}{2\cdot
|Te|}$ is used as a smoothing factor, where $Te$ denotes the test set.

The smoothed versions of $p(c_{j})$ and $\hat{p}(c_{j})$ are used
in place of their original versions in Equation \ref{eq:KLD}; as a
result, $KLD$ is always defined and still returns a value of 0 when
$p$ and $\hat{p}$ coincide.

$KLD$ is computed individually for each topic, and the results are averaged 
to yield the final score.


\subsection{Subtask E: Tweet quantification according to a five-point
scale}
\label{sec:4E}

\noindent Subtask E is an \emph{ordinal quantification} (OQ) task, in
which (as in OC) each tweet belongs exactly to one of the classes in
$\mathcal{C}$=\{\textsc{HighlyPositive}, \textsc{Positive},
\textsc{Neutral}, \textsc{Negative}, \textsc{HighlyNegative}\}, where
there is a total order on $\mathcal{C}$. As in binary quantification,
the task is to compute an estimate $\hat{p}(c_{j})$ of the relative
frequency $p(c_{j})$ in the test tweets of all the classes $c_{j}\in
\mathcal{C}$.

The measure we adopt for OQ is the \emph{Earth Mover's Distance}
\cite{Rubner:2000fk} (also known as the \emph{Vaser\u{s}te\u{\i}n
metric} \cite{Ruschendorf:2001mz}), a measure well-known in the field
of computer vision. $EMD$ is currently the only known measure for
ordinal quantification. It is defined for the general case in
which a distance $d(c',c'')$ is defined for each
$c',c''\in\mathcal{C}$. When there is a total order on the classes in
$\mathcal{C}$ and $d(c_i,c_{i+1})=1$ for all
$i\in\{1,...,(\mathcal{C}-1)\}$ (as in our
application), the Earth Mover's Distance is defined as
\begin{equation}
  \label{eq:EMD}
  EMD(\hat{p},p) = \sum_{j=1}^{|\mathcal{C}|-1}|\sum_{i=1}^{j}\hat{p}(c_{i})-\sum_{i=1}^{j}p(c_{i})|
\end{equation}
\noindent and can be computed in $|\mathcal{C}|$ steps from the
estimated and true class prevalences.

Like $KLD$ in Section \ref{sec:4D}, $EMD$ is a measure of error, so
lower values are better; $EMD$ ranges between 0 (best) and
$|\mathcal{C}|-1$ (worst). See \cite{Esuli:2010fk} for more details on
$EMD$.

As before, $EMD$ is computed individually for each topic, and the
results are then averaged across all topics to yield the final score.


\section{Participants and Results}
\label{sec:results}

\noindent A total of 43 teams (see Table \ref{tab:participants} at the
end of the paper) participated in SemEval-2016 Task 4, representing 25 countries;
the country with the highest participation was China (5 teams),
followed by Italy, Spain, and USA (4 teams each). The subtask with the
highest participation was Subtask A (34 teams), followed by Subtask B
(19 teams), Subtask D (14 teams), Subtask C (11 teams), and Subtask E
(10 teams).

It was not surprising that Subtask A proved to be the most popular --
it was a rerun from previous years; conversely, none among Subtasks B
to E had previously been offered in precisely the same
form. Quantification-related subtasks (D and E) generated 24
participations altogether, while subtasks with an ordinal nature (C
and E) attracted 21 participations. Only three teams participated in
all five subtasks; conversely, no less than 23 teams took part in one
subtask only (with a few exceptions, Subtask A).  Many teams that
participated in more than one subtask used essentially the same system
for all of them, with little tuning to the specifics of each subtask.

Few trends stand out among the participating systems. In terms of the
supervised learning methods used, there is a clear dominance of
methods based on deep learning, including convolutional neural
networks and recurrent neural networks (and, in particular, long
short-term memory networks); the software libraries for deep learning
most frequently used by the participants are Theano and
Keras. Conversely, kernel machines seem to be less frequently used
than in the past, and the use of learning methods other than the ones
mentioned above is scarce.

The use of distant supervision is ubiquitous; this is natural, since
there is an abundance of freely available tweets labelled according to
sentiment (possibly with silver labels only, e.g., emoticons), and it
is intuitive that their use as additional training data could be
helpful. Another ubiquitous technique is the use of word embeddings,
usually generated via either word2vec \cite{Mikolov:2013} or GloVe
\cite{pennington2014glove}; most authors seem to use general-purpose,
pre-trained embeddings, while some authors also use customized word
embeddings, trained either on the Tweet 2016 dataset or on tweet
datasets of some sort.

Nothing radically new seems to have emerged with respect to text
preprocessing; as in previous editions of this task, participants use
a mix of by now obvious techniques, such as negation scope detection,
elongation normalization, detection of amplifiers and diminishers,
plus the usual extraction of word $n$-grams, character $n$-grams, and
POS $n$-grams. The use of sentiment lexicons (alone or in combination
with each other; general-purpose or Twitter-specific) is obviously
still frequent.

In the next five subsections, we discuss the results of the
participating systems in the five subtasks, focusing on the techniques
and tools that the top-ranked participants have used. We also focus on
how the participants tailored (if at all) their approach to the
specific subtask. When discussing a specific subtask, we will adopt
the convention of adding to a team name a subscript which indicates
the position in the ranking for that subtask that the team obtained;
e.g., when discussing Subtask E, ``Finki$_{2}$'' indicates team
``Finki, which placed 2nd in the ranking for Subtask E''. The papers
describing the participants' approach are quoted in Table
\ref{tab:participants}.


\subsection{Subtask A: Message polarity classification}
\label{sec:4Adiscussion}

\noindent Table \ref{table:ResultsSubtaskA} ranks the systems submitted by the 34 teams who
participated in Subtask A ``Message Polarity Classification'' in terms
of the official measure $F_1^{PN}$. 
We further show the result for two other measures,
$\rho^{PN}$ (the measure that we adopted for Subtask B) and accuracy
($Acc=\frac{TP+TN}{TP+TN+FP+FN}$). We also report the result for a
baseline classifier that assigns to each tweet the \textsc{Positive}
class. For Subtask A evaluated using $F_1^{PN}$, this is the equivalent
of the majority class classifier for (binary or SLMC) classification
evaluated via vanilla accuracy, i.e., this is the ``smartest'' among
the trivial policies that attempt to maximize $F_1^{PN}$.

\renewcommand{\baselinestretch}{0.78}
\begin{table}[tbh]
  \centering
  \begin{small}
    \renewcommand{\arraystretch}{1.0}
    \resizebox{\columnwidth}{!} {
    \begin{tabular}{|c|l|l|l|l|}
      \hline\rule[0ex]{0mm}{2ex}
      \bf \# & \multicolumn{1}{c|}{\bf System} & \multicolumn{1}{c|}{$F_{1}^{PN}$} & \multicolumn{1}{c|}{$\rho^{PN}$} & \multicolumn{1}{c|}{$Acc$} \\
      \hline\rule[0ex]{-1mm}{2ex} 
      \bf 1 & SwissCheese & \bf 0.633$_{1}$ & 0.667$_{2}$ & 0.646$_{1}$ \\
      \bf 2 & SENSEI-LIF & \bf 0.630$_{2}$ & 0.670$_{1}$ & 0.617$_{7}$ \\
      \bf 3 & UNIMELB & \bf 0.617$_{3}$ & 0.641$_{5}$ & 0.616$_{8}$ \\
      \bf 4 & INESC-ID & \bf 0.610$_{4}$ & 0.662$_{3}$ & 0.600$_{10}$ \\
      \bf 5 & aueb.twitter.sentiment & \bf 0.605$_{5}$ & 0.644$_{4}$ & 0.629$_{6}$ \\
      \bf 6 & SentiSys & \bf 0.598$_{6}$ & 0.641$_{5}$ & 0.609$_{9}$ \\
      \bf 7 & I2RNTU & \bf 0.596$_{7}$ & 0.637$_{7}$ & 0.593$_{12}$ \\
      \bf 8 & INSIGHT-1 & \bf 0.593$_{8}$ & 0.616$_{11}$ & 0.635$_{5}$ \\
      \bf 9 & TwiSE & \bf 0.586$_{9}$ & 0.598$_{16}$ & 0.528$_{24}$ \\
      \bf 10 & ECNU (*) & \bf 0.585$_{10}$ & 0.617$_{10}$ & 0.571$_{16}$ \\
      \bf 11 & NTNUSentEval & \bf 0.583$_{11}$ & 0.619$_{8}$ & 0.643$_{2}$ \\
      \bf 12 & MDSENT & \bf 0.580$_{12}$ & 0.592$_{18}$ & 0.545$_{20}$ \\
      & CUFE & \bf 0.580$_{12}$ & 0.619$_{8}$ & 0.637$_{4}$ \\
      \bf 14 & THUIR & \bf 0.576$_{14}$ & 0.605$_{15}$ & 0.596$_{11}$ \\
      & PUT & \bf 0.576$_{14}$ & 0.607$_{13}$ & 0.584$_{14}$ \\
      \bf 16 & LYS & \bf 0.575$_{16}$ & 0.615$_{12}$ & 0.585$_{13}$ \\
      \bf 17 & IIP & \bf 0.574$_{17}$ & 0.579$_{19}$ & 0.537$_{23}$ \\
      \bf 18 & UniPI & \bf 0.571$_{18}$ & 0.607$_{13}$ & 0.639$_{3}$ \\
      \bf 19 & DIEGOLab16 (*) & \bf 0.554$_{19}$ & 0.593$_{17}$ & 0.549$_{19}$ \\
      \bf 20 & GTI & \bf 0.539$_{20}$ & 0.557$_{21}$ & 0.518$_{26}$ \\
      \bf 21 & OPAL & \bf 0.505$_{21}$ & 0.560$_{20}$ & 0.541$_{22}$ \\
      \bf 22 & DSIC-ELIRF & \bf 0.502$_{22}$ & 0.511$_{25}$ & 0.513$_{27}$ \\
      \bf 23 & UofL & \bf 0.499$_{23}$ & 0.537$_{22}$ & 0.572$_{15}$ \\
      & ELiRF & \bf 0.499$_{23}$ & 0.516$_{24}$ & 0.543$_{21}$ \\
      \bf 25 & ISTI-CNR & \bf 0.494$_{25}$ & 0.529$_{23}$ & 0.567$_{17}$ \\
      \bf 26 & SteM & \bf 0.478$_{26}$ & 0.496$_{27}$ & 0.452$_{31}$ \\
      \bf 27 & Tweester & \bf 0.455$_{27}$ & 0.503$_{26}$ & 0.523$_{25}$ \\
      \bf 28 & Minions & \bf 0.415$_{28}$ & 0.485$_{28}$ & 0.556$_{18}$ \\
      \bf 29 & Aicyber & \bf 0.402$_{29}$ & 0.457$_{29}$ & 0.506$_{28}$ \\
      \bf 30 & mib & \bf 0.401$_{30}$ & 0.438$_{30}$ & 0.480$_{29}$ \\
      \bf 31 & VCU-TSA & \bf 0.372$_{31}$ & 0.390$_{32}$ & 0.382$_{32}$ \\
      \bf 32 & SentimentalITists & \bf 0.339$_{32}$ & 0.424$_{31}$ & 0.480$_{29}$ \\
      \bf 33 & WR & \bf 0.330$_{33}$ & 0.333$_{34}$ & 0.298$_{34}$ \\
      \bf 34 & CICBUAPnlp & \bf 0.303$_{34}$ & 0.377$_{33}$ & 0.374$_{33}$ \\
      \hline\rule[0ex]{0mm}{2ex} 
      & Baseline & \bf 0.255 & 0.333 & 0.342 \\ 
      \hline
    \end{tabular}
    }
\caption{\label{table:ResultsSubtaskA} Results for Subtask A ``Message Polarity Classification'' on the Tweet 2016 dataset. The systems are ordered by their $F_1^{PN}$ score. In each column the rankings according to the corresponding measure are indicated with a subscript. Teams marked as ``(*)'' are late submitters, i.e., their original submission was deemed irregular by the organizers, and a revised submission was entered after the deadline.
}
\end{small}
\end{table}
\renewcommand{\baselinestretch}{1.00}

All 34 participating systems were able to outperform
the baseline on all three measures, with the exception of one system that scored below the baseline on $Acc$. The
top-scoring team (SwissCheese$_{1}$) used an ensemble of convolutional
neural networks, differing in their choice of filter shapes, pooling
shapes and usage of hidden layers. Word embeddings generated via
word2vec were also used, and the neural networks were trained by using
distant supervision. Out of
the 10 top-ranked teams, 5 teams (SwissCheese$_{1}$, SENSEI-LIF$_{2}$,
UNIMELB$_{3}$, INESC-ID$_{4}$, INSIGHT-1$_{8}$) used deep NNs of some
sort, and 7 teams (SwissCheese$_{1}$, SENSEI-LIF$_{2}$, UNIMELB$_{3}$,
INESC-ID$_{4}$, aueb.twitter.sentiment$_{5}$, I2RNTU$_{7}$,
INSIGHT-1$_{8}$) used either general-purpose or task-specific word
embeddings, generated via word2vec or GloVe.

\paragraph{Historical results.} We also tested the participating
systems on the test sets from the three previous editions of this
subtask. Participants were not allowed to use these test sets for
training. Results (measured on $F_1^{PN}$) are
reported in Table~\ref{table:ResultsHistoricalSubtaskA}. The top-performing systems on Tweet 2016 are also top-ranked on
the test datasets from previous years. There is a general
pattern: the top-ranked system in year $x$
outperforms the top-ranked system in year $(x-1)$ on the official
dataset of year $(x-1)$. Top-ranked systems tend to
use approaches that are universally strong, even when tested on
out-of-domain test sets such as SMS, LiveJournal, or sarcastic tweets
(yet, for sarcastic tweets, there are larger differences in rank
compared to systems rankings on Tweet 2016).  It is unclear where
improvements come from: (a)~the additional training data
that we made available this year (in addition to Tweet-train-2013,
which was used in 2013--2015), thus effectively doubling the amount of
training data, or (b)~because of advancement of learning methods.

\renewcommand{\baselinestretch}{0.78}
\begin{table*}[thb!]
  \centering
  \begin{small}
    \renewcommand{\arraystretch}{1.0}
    \resizebox{\textwidth}{!} {
    \begin{tabular}{|c|l|ll|lll|l|l|}
      \hline\rule[0ex]{0mm}{2ex} 
      & & \multicolumn{2}{c|}{\bf 2013} & \multicolumn{3}{c|}{\bf 2014} & \multicolumn{1}{c|}{\bf 2015} & \multicolumn{1}{c|}{\bf 2016} \\
      \bf \# & \multicolumn{1}{c|}{\bf System} & \multicolumn{1}{c}{\bf Tweet} & \multicolumn{1}{c|}{\bf SMS} & \multicolumn{1}{c}{\bf Tweet} & \multicolumn{1}{c}{\bf Tweet} & \multicolumn{1}{c|}{\bf Live-} & \multicolumn{1}{c|}{\bf Tweet} & \multicolumn{1}{c|}{\bf Tweet} \\
      & & & & \bf & \bf sarcasm & \bf Journal & & \bf \\
      \hline\rule[0ex]{-1mm}{2ex} 
      \bf 1 & SwissCheese & 0.700$_{4}$ & 0.637$_{2}$ & 0.716$_{4}$ & 0.566$_{1}$ & 0.695$_{7}$ & 0.671$_{1}$ & \bf 0.633$_{1}$ \\
      \bf 2 & SENSEI-LIF & 0.706$_{3}$ & 0.634$_{3}$ & 0.744$_{1}$ & 0.467$_{8}$ & 0.741$_{1}$ & 0.662$_{2}$ & \bf 0.630$_{2}$ \\
      \bf 3 & UNIMELB & 0.687$_{6}$ & 0.593$_{9}$ & 0.706$_{6}$ & 0.449$_{11}$ & 0.683$_{9}$ & 0.651$_{4}$ & \bf 0.617$_{3}$ \\
      \bf 4 & INESC-ID & 0.723$_{1}$ & 0.609$_{6}$ & 0.727$_{2}$ & 0.554$_{2}$ & 0.702$_{4}$ & 0.657$_{3}$ & \bf 0.610$_{4}$ \\
      \bf 5 & aueb.twitter.sentiment & 0.666$_{7}$ & 0.618$_{5}$ & 0.708$_{5}$ & 0.410$_{17}$ & 0.695$_{7}$ & 0.623$_{7}$ & \bf 0.605$_{5}$ \\
      \bf 6 & SentiSys & 0.714$_{2}$ & 0.633$_{4}$ & 0.723$_{3}$ & 0.515$_{4}$ & 0.726$_{2}$ & 0.644$_{5}$ & \bf 0.598$_{6}$ \\
      \bf 7 & I2RNTU & 0.693$_{5}$ & 0.597$_{7}$ & 0.680$_{7}$ & 0.469$_{6}$ & 0.696$_{6}$ & 0.638$_{6}$ & \bf 0.596$_{7}$ \\
      \bf 8 & INSIGHT-1 & 0.602$_{16}$ & 0.582$_{12}$ & 0.644$_{15}$ & 0.391$_{23}$ & 0.559$_{23}$ & 0.595$_{16}$ & \bf 0.593$_{8}$ \\
      \bf 9 & TwiSE & 0.610$_{15}$ & 0.540$_{16}$ & 0.645$_{13}$ & 0.450$_{10}$ & 0.649$_{13}$ & 0.621$_{8}$ & \bf 0.586$_{9}$ \\
      \bf 10 & ECNU (*) & 0.643$_{9}$ & 0.593$_{9}$ & 0.662$_{8}$ & 0.425$_{14}$ & 0.663$_{10}$ & 0.606$_{11}$ & \bf 0.585$_{10}$ \\
      \bf 11 & NTNUSentEval & 0.623$_{11}$ & 0.641$_{1}$ & 0.651$_{10}$ & 0.427$_{13}$ & 0.719$_{3}$ & 0.599$_{13}$ & \bf 0.583$_{11}$ \\
      \bf 12 & MDSENT & 0.589$_{19}$ & 0.509$_{20}$ & 0.587$_{20}$ & 0.386$_{24}$ & 0.606$_{18}$ & 0.593$_{17}$ & \bf 0.580$_{12}$ \\
      & CUFE & 0.642$_{10}$ & 0.596$_{8}$ & 0.662$_{8}$ & 0.466$_{9}$ & 0.697$_{5}$ & 0.598$_{14}$ & \bf 0.580$_{12}$ \\
      \bf 14 & THUIR & 0.616$_{12}$ & 0.575$_{14}$ & 0.648$_{11}$ & 0.399$_{20}$ & 0.640$_{15}$ & 0.617$_{10}$ & \bf 0.576$_{14}$ \\
      & PUT & 0.565$_{21}$ & 0.511$_{19}$ & 0.614$_{19}$ & 0.360$_{27}$ & 0.648$_{14}$ & 0.597$_{15}$ & \bf 0.576$_{14}$ \\
      \bf 16 & LYS & 0.650$_{8}$ & 0.579$_{13}$ & 0.647$_{12}$ & 0.407$_{18}$ & 0.655$_{11}$ & 0.603$_{12}$ & \bf 0.575$_{16}$ \\
      \bf 17 & IIP & 0.598$_{17}$ & 0.465$_{23}$ & 0.645$_{13}$ & 0.405$_{19}$ & 0.640$_{15}$ & 0.619$_{9}$ & \bf 0.574$_{17}$ \\
      \bf 18 & UniPI & 0.592$_{18}$ & 0.585$_{11}$ & 0.627$_{17}$ & 0.381$_{25}$ & 0.654$_{12}$ & 0.586$_{18}$ & \bf 0.571$_{18}$ \\
      \bf 19 & DIEGOLab16 (*) & 0.611$_{14}$ & 0.506$_{21}$ & 0.618$_{18}$ & 0.497$_{5}$ & 0.594$_{20}$ & 0.584$_{19}$ & \bf 0.554$_{19}$ \\
      \bf 20 & GTI & 0.612$_{13}$ & 0.524$_{17}$ & 0.639$_{16}$ & 0.468$_{7}$ & 0.623$_{17}$ & 0.584$_{19}$ & \bf 0.539$_{20}$ \\
      \bf 21 & OPAL & 0.567$_{20}$ & 0.562$_{15}$ & 0.556$_{23}$ & 0.395$_{21}$ & 0.593$_{21}$ & 0.531$_{21}$ & \bf 0.505$_{21}$ \\
      \bf 22 & DSIC-ELIRF & 0.494$_{25}$ & 0.404$_{26}$ & 0.546$_{26}$ & 0.342$_{29}$ & 0.517$_{24}$ & 0.531$_{21}$ & \bf 0.502$_{22}$ \\
      \bf 23 & UofL & 0.490$_{26}$ & 0.443$_{24}$ & 0.547$_{25}$ & 0.372$_{26}$ & 0.574$_{22}$ & 0.502$_{25}$ & \bf 0.499$_{23}$ \\
      & ELiRF & 0.462$_{28}$ & 0.408$_{25}$ & 0.514$_{28}$ & 0.310$_{33}$ & 0.493$_{25}$ & 0.493$_{26}$ & \bf 0.499$_{23}$ \\
      \bf 25 & ISTI-CNR & 0.538$_{22}$ & 0.492$_{22}$ & 0.572$_{21}$ & 0.327$_{30}$ & 0.598$_{19}$ & 0.508$_{24}$ & \bf 0.494$_{25}$ \\
      \bf 26 & SteM & 0.518$_{23}$ & 0.315$_{29}$ & 0.571$_{22}$ & 0.320$_{32}$ & 0.405$_{28}$ & 0.517$_{23}$ & \bf 0.478$_{26}$ \\
      \bf 27 & Tweester & 0.506$_{24}$ & 0.340$_{28}$ & 0.529$_{27}$ & 0.540$_{3}$ & 0.379$_{29}$ & 0.479$_{28}$ & \bf 0.455$_{27}$ \\
      \bf 28 & Minions & 0.489$_{27}$ & 0.521$_{18}$ & 0.554$_{24}$ & 0.420$_{16}$ & 0.475$_{26}$ & 0.481$_{27}$ & \bf 0.415$_{28}$ \\
      \bf 29 & Aicyber & 0.418$_{29}$ & 0.361$_{27}$ & 0.457$_{29}$ & 0.326$_{31}$ & 0.440$_{27}$ & 0.432$_{29}$ & \bf 0.402$_{29}$ \\
      \bf 30 & mib & 0.394$_{30}$ & 0.310$_{30}$ & 0.415$_{31}$ & 0.352$_{28}$ & 0.359$_{31}$ & 0.413$_{31}$ & \bf 0.401$_{30}$ \\
      \bf 31 & VCU-TSA & 0.383$_{31}$ & 0.307$_{31}$ & 0.444$_{30}$ & 0.425$_{14}$ & 0.336$_{32}$ & 0.416$_{30}$ & \bf 0.372$_{31}$ \\
      \bf 32 & SentimentalITists & 0.339$_{33}$ & 0.238$_{33}$ & 0.393$_{32}$ & 0.288$_{34}$ & 0.323$_{34}$ & 0.343$_{33}$ & \bf 0.339$_{32}$ \\
      \bf 33 & WR & 0.355$_{32}$ & 0.284$_{32}$ & 0.393$_{32}$ & 0.430$_{12}$ & 0.366$_{30}$ & 0.377$_{32}$ & \bf 0.330$_{33}$ \\
      \bf 34 & CICBUAPnlp & 0.193$_{34}$ & 0.193$_{34}$ & 0.335$_{34}$ & 0.393$_{22}$ & 0.326$_{33}$ & 0.303$_{34}$ & \bf 0.303$_{34}$ \\
      \hline
    \end{tabular}\medskip
    }
    \caption{\label{table:ResultsHistoricalSubtaskA} Historical
    results for Subtask A ``Message Polarity Classification''. The
    systems are ordered by their score on the Tweet 2016 dataset; the
    rankings on the individual datasets are indicated with a
    subscript. The meaning of ``(*)'' is as in Table
    \ref{table:ResultsSubtaskA}.}
  \end{small}
\end{table*}
\renewcommand{\baselinestretch}{1.00}

We further look at the top scores achieved by any system in the period
2013--2016.  The results are shown in
Table~\ref{table:ResultsHistoricalSubtaskA:best}.  Interestingly, the
results for a test set improve in the second year it is used (i.e.,
the year after it was used as an official test set) by 1--3 points
absolute, but then do not improve further and stay stable,
or can even decrease a bit. This might be due to participants
optimizing their systems primarily on the test set from the preceding
year.

\begin{table*}[htb]
  \centering
  \begin{small}
    \renewcommand{\arraystretch}{1.0}
    \begin{tabular}{|c|cc|ccc|c|c|}
      \hline\rule[0ex]{0mm}{2ex} 
      & \multicolumn{2}{c|}{\bf 2013} & \multicolumn{3}{c|}{\bf 2014} & \multicolumn{1}{c|}{\bf 2015} & \multicolumn{1}{c|}{\bf 2016} \\
      \multicolumn{1}{|c|}{\bf Year} & \multicolumn{1}{c}{\bf Tweet} & \multicolumn{1}{c|}{\bf SMS} & \multicolumn{1}{c}{\bf Tweet} & \multicolumn{1}{c}{\bf Tweet} & \multicolumn{1}{c|}{\bf Live-} & \multicolumn{1}{c|}{\bf Tweet} & \multicolumn{1}{c|}{\bf Tweet} \\
      & & & \bf & \bf sarcasm & \bf Journal & & \bf \\
      \hline\rule[0ex]{-1mm}{2ex} 
      Best in 2016 & 0.723 & 0.641 & 0.744 & 0.566 & 0.741 & 0.671 & 0.633\\
      Best in 2015 & 0.728 & 0.685 & 0.744 & 0.591 & 0.753 & 0.648 & \multicolumn{1}{c|}{--} \\
      Best in 2014 & 0.721 & 0.703 & 0.710 & 0.582 & 0.748 & \multicolumn{1}{c|}{--} & \multicolumn{1}{c|}{--} \\
      Best in 2013 & 0.690 & 0.685 & \multicolumn{1}{c}{--} & \multicolumn{1}{c}{--} & \multicolumn{1}{c|}{--} & \multicolumn{1}{c|}{--} & \multicolumn{1}{c|}{--}\\
      \hline
    \end{tabular}\medskip
    \caption{\label{table:ResultsHistoricalSubtaskA:best} Historical
    results for the best systems for Subtask A ``Message Polarity
    Classification'' over the years 2013--2016.}
  \end{small}
\end{table*}
\renewcommand{\baselinestretch}{1.00}


\subsection{Subtask B: Tweet classification according to a two-point
scale}
\label{sec:4Bdiscussion}

\noindent Table \ref{table:ResultsSubtaskB} ranks the 19 teams who
participated in Subtask B ``Tweet classification according to a
two-point scale'' in terms of the official measure $\rho^{PN}$. Two
other measures are reported, $F_{1}^{PN}$ (the measure adopted for
Subtask A) and accuracy ($Acc$). We also report the result of a baseline
that assigns to each tweet the positive class. This is the
``smartest'' among the trivial policies that attempt to maximize
$\rho^{PN}$. This baseline always returns $\rho^{PN}=0.500$.

Note however that this is also (\emph{i}) the value returned by the
classifier that assigns to each tweet the negative class, and
(\emph{ii}) the expected value returned by the random classifier; for
more details see \cite[Section 5]{Sebastiani:2015zl}, where
$\rho^{PN}$ is called $K$.

\renewcommand{\baselinestretch}{0.78}
\begin{table}[htb]
  \centering
  \begin{small}
    \renewcommand{\arraystretch}{1.0}
    \resizebox{\columnwidth}{!} {
    \begin{tabular}{|c|l|l|l|l|}
      \hline\rule[0ex]{0mm}{2ex} 
      \bf \# & \bf System & \multicolumn{1}{c|}{$\rho^{PN}$} & \multicolumn{1}{c|}{$F_{1}^{PN}$} & \multicolumn{1}{c|}{$Acc$} \\
      \hline\rule[0ex]{-1mm}{2ex} 
      \bf 1 & Tweester & \bf 0.797$_{1}$ & 0.799$_{1}$ & 0.862$_{3}$ \\
      \bf 2 & LYS & \bf 0.791$_{2}$ & 0.720$_{10}$ & 0.762$_{17}$ \\
      \bf 3 & thecerealkiller & \bf 0.784$_{3}$ & 0.762$_{5}$ & 0.823$_{9}$ \\
      \bf 4 & ECNU (*) & \bf 0.768$_{4}$ & 0.770$_{4}$ & 0.843$_{5}$ \\
      \bf 5 & INSIGHT-1 & \bf 0.767$_{5}$ & 0.786$_{3}$ & 0.864$_{2}$ \\
      \bf 6 & PUT & \bf 0.763$_{6}$ & 0.732$_{8}$ & 0.794$_{14}$ \\
      \bf 7 & UNIMELB & \bf 0.758$_{7}$ & 0.788$_{2}$ & 0.870$_{1}$ \\
      \bf 8 & TwiSE & \bf 0.756$_{8}$ & 0.752$_{6}$ & 0.826$_{8}$ \\
      \bf 9 & GTI & \bf 0.736$_{9}$ & 0.731$_{9}$ & 0.811$_{11}$ \\
      \bf 10 & Finki & \bf 0.720$_{10}$ & 0.748$_{7}$ & 0.848$_{4}$ \\
      \bf 11 & pkudblab & \bf 0.689$_{11}$ & 0.716$_{11}$ & 0.832$_{7}$ \\
      \bf 12 & CUFE & \bf 0.679$_{12}$ & 0.708$_{12}$ & 0.834$_{6}$ \\
      \bf 13 & ISTI-CNR & \bf 0.671$_{13}$ & 0.690$_{13}$ & 0.811$_{11}$ \\
      \bf 14 & SwissCheese & \bf 0.648$_{14}$ & 0.674$_{14}$ & 0.820$_{10}$ \\
      \bf 15 & SentimentalITists & \bf 0.624$_{15}$ & 0.643$_{15}$ & 0.802$_{13}$ \\
      \bf 16 & PotTS & \bf 0.618$_{16}$ & 0.610$_{17}$ & 0.712$_{18}$ \\
      \bf 17 & OPAL & \bf 0.616$_{17}$ & 0.633$_{16}$ & 0.792$_{15}$ \\
      \bf 18 & WR & \bf 0.522$_{18}$ & 0.502$_{18}$ & 0.577$_{19}$ \\
      \bf 19 & VCU-TSA & \bf 0.502$_{19}$ & 0.448$_{19}$ & 0.775$_{16}$ \\
      \hline\rule[0ex]{0mm}{2ex} 
      & Baseline & \bf 0.500 & 0.438 & 0.778\\
      \hline
    \end{tabular}
    } \medskip
 \caption{\label{table:ResultsSubtaskB} Results for Subtask B ``Tweet classification according to a two-point scale'' on the Tweet 2016 dataset. The systems are ordered by their $\rho^{PN} $ score (higher is better). The meaning of ``(*)'' is as in Table \ref{table:ResultsSubtaskA}.
 }
\end{small}
\end{table}
\renewcommand{\baselinestretch}{1.00}

The top-scoring team (Tweester$_{1}$) used a combination of
convolutional neural networks, topic modeling, and word embeddings
generated via word2vec. Similar to Subtask A, the main trend among all
participants is the widespread use of deep learning techniques. 

Out of the 10 top-ranked participating teams, 5 teams (Tweester$_{1}$, LYS$_{2}$,
INSIGHT-1$_{5}$, UNIMELB$_{7}$, Finki$_{10}$) used convolutional neural networks;
3 teams (thecerealkiller$_{3}$, UNIMELB$_{7}$, Finki$_{10}$) submitted systems using
recurrent neural networks; and 7 teams (Tweester$_{1}$, LYS$_{2}$,
INSIGHT-1$_{5}$, UNIMELB$_{7}$, Finki$_{10}$) incorporated in their participating systems either
general-purpose or task-specific word embeddings (generated via toolkits such as GloVe or word2vec). 

Conversely, the use of classifiers such as support vector machines, which were dominant until a
few years ago, seems to have decreased, with only one team (TwiSE$_{8}$)
in the top 10 using them.


\subsection{Subtask C: Tweet classification according to a five-point
scale}
\label{sec:4Cdiscussion}

\noindent Table \ref{table:ResultsSubtaskC} ranks the 11 teams who
participated in Subtask C ``Tweet classification according to a
five-point scale'' in terms of the official measure $MAE^{M}$; we also show $MAE^{\mu}$ (see Equation \ref{eq:microMAE}). We
also report the result of a baseline system that assigns to each tweet
the middle class (i.e., \textsc{Neutral}); for ordinal
classification evaluated via $MAE^M$, this is 
the majority-class classifier for (binary or SLMC) classification
evaluated via vanilla accuracy, i.e., this is
\cite{Baccianella:2009qd} the ``smartest'' among the trivial policies
that attempt to maximize $MAE^M$.

\renewcommand{\baselinestretch}{0.78}
\begin{table}[htb]
  \centering
  \begin{small}
    \renewcommand{\arraystretch}{1.0}
    \resizebox{\columnwidth}{!} {
    \begin{tabular}{|c|l|l|l|}
      \hline\rule[0ex]{0mm}{2ex} 
      \bf \# & \bf System & \multicolumn{1}{c|}{$MAE^M$} & \multicolumn{1}{c|}{$MAE^{\mu}$} \\
      \hline\rule[0ex]{-1mm}{2ex} 
      \bf 1 & TwiSE & \bf 0.719$_{1}$ & 0.632$_{5}$ \\
      \bf 2 & ECNU (*) & \bf 0.806$_{2}$ & 0.726$_{8}$ \\
      \bf 3 & PUT & \bf 0.860$_{3}$ & 0.773$_{9}$ \\
      \bf 4 & LYS & \bf 0.864$_{4}$ & 0.694$_{7}$ \\
      \bf 5 & Finki & \bf 0.869$_{5}$ & 0.672$_{6}$ \\
      \bf 6 & INSIGHT-1 & \bf 1.006$_{6}$ & 0.607$_{3}$ \\
      \bf 7 & ISTI-CNR & \bf 1.074$_{7}$ & 0.580$_{1}$ \\
      \bf 8 & YZU-NLP & \bf 1.111$_{8}$ & 0.588$_{2}$ \\
      \bf 9 & SentimentalITists & \bf 1.148$_{9}$ & 0.625$_{4}$ \\
      \bf 10 & PotTS & \bf 1.237$_{10}$ & 0.860$_{10}$ \\
      \bf 11 & pkudblab & \bf 1.697$_{11}$ & 1.300$_{11}$ \\
      \hline\rule[0ex]{0mm}{2ex} 
      & Baseline & \bf 1.200 & 0.537\\
      \hline
    \end{tabular}
    } \medskip
 \caption{\label{table:ResultsSubtaskC} Results for Subtask C ``Tweet classification according to a five-point scale'' on the Tweet 2016 dataset. The systems are ordered by their $MAE^{M}$ score (lower is better). The meaning of ``(*)'' is as in Table \ref{table:ResultsSubtaskA}.
 }
\end{small}
\end{table}
\renewcommand{\baselinestretch}{1.00}

The top-scoring team (TwiSE$_{1}$) used a single-label multi-class classifier
to classify the tweets according to their overall polarity. 
In particular, they used logistic regression that minimizes the multinomial loss across the classes, with weights to cope with class imbalance. Note that they ignored the given topics altogether.


\renewcommand{\baselinestretch}{0.78}
\begin{table}[tbh]
  \centering
  \begin{small}
    \renewcommand{\arraystretch}{1.0}
    \resizebox{\columnwidth}{!} {
    \begin{tabular}{|c|l|l|l|l|}
      \hline\rule[0ex]{0mm}{2ex}
      \bf \# & \bf System & \multicolumn{1}{c|}{$KLD$} & \multicolumn{1}{c|}{$AE$} &\multicolumn{1}{c|}{$RAE$} \\
      \hline\rule[0ex]{-1mm}{2ex}
      \bf 1 & Finki & \bf 0.034$_{1}$ & 0.074$_{1}$ & 0.707$_{3}$ \\
      \bf 2 & LYS & \bf 0.053$_{2}$ & 0.099$_{4}$ & 0.844$_{5}$ \\
      & TwiSE & \bf 0.053$_{2}$ & 0.101$_{5}$ & 0.864$_{6}$ \\
      \bf 4 & INSIGHT-1 & \bf 0.054$_{4}$ & 0.085$_{2}$ & 0.423$_{1}$ \\
      \bf 5 & GTI & \bf 0.055$_{5}$ & 0.104$_{6}$ & 1.200$_{10}$ \\
      & QCRI & \bf 0.055$_{5}$ & 0.095$_{3}$ & 0.864$_{6}$ \\
      \bf 7 & NRU-HSE & \bf 0.084$_{7}$ & 0.120$_{8}$ & 0.767$_{4}$ \\
      \bf 8 & PotTS & \bf 0.094$_{8}$ & 0.150$_{12}$ & 1.838$_{12}$ \\
      \bf 9 & pkudblab & \bf 0.099$_{9}$ & 0.109$_{7}$ & 0.947$_{8}$ \\
      \bf 10 & ECNU (*) & \bf 0.121$_{10}$ & 0.148$_{11}$ & 1.171$_{9}$ \\
      \bf 11 & ISTI-CNR & \bf 0.127$_{11}$ & 0.147$_{9}$ & 1.371$_{11}$ \\
      \bf 12 & SwissCheese & \bf 0.191$_{12}$ & 0.147$_{9}$ & 0.638$_{2}$ \\
      \bf 13 & UDLAP & \bf 0.261$_{13}$ & 0.274$_{13}$ & 2.973$_{13}$ \\
      \bf 14 & HSENN & \bf 0.399$_{14}$ & 0.336$_{14}$ & 3.930$_{14}$ \\
      \hline\rule[0ex]{0mm}{2ex} 
      & Baseline$_1$ & \bf 0.175 & 0.184 & 2.110 \\
      & Baseline$_2$ & \bf 0.887 & 0.242 & 1.155 \\
      \hline
    \end{tabular}
    }
 \caption{\label{table:ResultsSubtaskD} Results for Subtask D ``Tweet quantification according to a two-point scale'' on the Tweet 2016 dataset. The systems are ordered by their $KLD$ score (lower is better). The meaning of ``(*)'' is as in Table \ref{table:ResultsSubtaskA}.
 }
\end{small}
\end{table}
\renewcommand{\baselinestretch}{1.00}

Only 2 of the 11 participating teams tuned their systems to exploit
the ordinal (as opposed to binary, or single-label multi-class) nature
of this subtask. The two teams who did exploit the ordinal nature of
the problem are PUT$_{3}$, which uses an ensemble of ordinal
regression approaches, and ISTI-CNR$_{7}$, which uses a
tree-based approach to ordinal regression. All other teams used
general-purpose approaches for single-label multi-class
classification, in many cases relying (as for Subtask B) on
convolutional neural networks, recurrent neural networks, and word embeddings.


\subsection{Subtask D: Tweet quantification according to a two-point
scale}
\label{sec:4Ddiscussion}

\noindent Table \ref{table:ResultsSubtaskD} ranks the 14 teams who
participated in Subtask D ``Tweet quantification according to a
two-point scale'' on the official measure $KLD$. Two other
measures are reported, \emph{absolute error} ($AE$):
\begin{equation}\label{eq:AE}
  AE(p,\hat{p},\mathcal{C})=\frac{1}{|\mathcal{C}|}\sum_{c\in \mathcal{C}}|\hat{p}(c)-p(c)|
\end{equation}
\noindent and \emph{relative absolute error} ($RAE$):
\begin{equation}\label{eq:rae}
  RAE(p,\hat{p},\mathcal{C})=\frac{1}{|\mathcal{C}|}\sum_{c\in 
  \mathcal{C}}\displaystyle\frac{|\hat{p}(c)-p(c)|}{p(c)}
\end{equation}
\noindent where the notation is the same as in Equation
\ref{eq:KLD}. 

We also report the result of a ``maximum likelihood''
baseline system (dubbed Baseline$_1$). This system assigns to each
test topic the distribution of the training tweets (the union of
TRAIN, DEV, DEVTEST) across the classes. This is the ``smartest''
among the trivial policies that attempt to maximize $KLD$. We also
report the result of a further (less smart) baseline system (dubbed
Baseline$_2$), i.e., one that assigns a prevalence of 1 to the
majority class (which happens to be the \textsc{Positive} class) and a
prevalence of 0 to the other class.

The top-scoring team (Finki$_{1}$) adopts an approach based on
``classify and count'', a classification-oriented (instead of quantification-oriented) approach, using recurrent and convolutional
neural networks, and GloVe word embeddings.

Indeed, only 5 of the 14 participating teams tuned their systems to
the fact that it deals with
quantification (as opposed to classification). Among the teams who do
rely on quantification-oriented approaches, teams LYS$_{2}$ and
HSENN$_{14}$ used an existing structured prediction method that
directly optimizes $KLD$; teams QCRI$_{5}$ and ISTI-CNR$_{11}$ use
existing probabilistic quantification methods; team NRU-HSE$_{7}$ uses
an existing iterative quantification method based on cost-sensitive
learning. Interestingly, team TwiSE$_{2}$ uses a ``classify and
count'' approach after comparing it with a quantification-oriented
method (similar to the one used by teams LYS$_{2}$ and HSENN$_{14}$)
on the development set, and concluding that the former works better
than the latter. All other teams used ``classify and count''
approaches, mostly based on convolutional neural networks and word
embeddings.


\subsection{Subtask E: Tweet quantification according to a five-point
scale}
\label{sec:4Ediscussion}

\noindent Table \ref{table:ResultsSubtaskE} lists the results obtained
by the 10 participating teams on Subtask E ``Tweet quantification according
to a five-point scale''. We also report the result of a ``maximum
likelihood'' baseline system (dubbed Baseline$_1$), i.e., one that
assigns to each test topic the same distribution, namely the distribution of the training tweets
(the union of TRAIN, DEV, DEVTEST) across the classes; this is the
``smartest'' among the trivial policies (i.e., those that do not
require any genuine work) that attempt to maximize $EMD$.

We further report the result of less smart baseline system (dubbed
Baseline$_2$) -- one that assigns a prevalence of 1 to the
majority class (which coincides with the \textsc{Positive} class) and a
prevalence of 0 to all other classes.

\renewcommand{\baselinestretch}{0.78}
\begin{table}[htb]
  \centering
  \begin{small}
    \renewcommand{\arraystretch}{1.0}
    \begin{tabular}{|c|l|l|}
      \hline\rule[0ex]{0mm}{2ex} 
      \bf \# & \bf System & \multicolumn{1}{c|}{$EMD$} \\
      \hline\rule[0ex]{-1mm}{2ex}
      \bf 1 & QCRI & \bf 0.243$_{1}$ \\
      \bf 2 & Finki & \bf 0.316$_{2}$ \\
      \bf 3 & pkudblab & \bf 0.331$_{3}$ \\
      \bf 4 & NRU-HSE & \bf 0.334$_{4}$ \\
      \bf 5 & ECNU (*) & \bf 0.341$_{5}$ \\
      \bf 6 & ISTI-CNR & \bf 0.358$_{6}$ \\
      \bf 7 & LYS & \bf 0.360$_{7}$ \\
      \bf 8 & INSIGHT-1 & \bf 0.366$_{8}$ \\
      \bf 9 & HSENN & \bf 0.545$_{9}$ \\
      \bf 10 & PotTS & \bf 0.818$_{10}$ \\
      \hline\rule[0ex]{0mm}{2ex} 
      & Baseline$_1$ & \bf 0.474 \\
      & Baseline$_2$ & \bf 0.734 \\
      \hline
    \end{tabular}\medskip
 \caption{\label{table:ResultsSubtaskE} Results for Subtask E ``Tweet quantification according to a five-point scale'' on the Tweet 2016 dataset. The systems are ordered by their $EMD$ score (lower is better). The meaning of ``(*)'' is as in Table \ref{table:ResultsSubtaskA}.
 }
\end{small}
\end{table}
\renewcommand{\baselinestretch}{1.00}

Only 3 of the 10 participants tuned their systems to the specific
characteristics of this subtask, i.e., to the fact that it deals with
quantification (as opposed to classification) \textit{and} to the fact
that it has an ordinal (as opposed to binary) nature. 

In particular, the top-scoring
team (QCRI$_{1}$) used a novel algorithm explicitly designed for
ordinal quantification, that leverages an ordinal hierarchy of binary
probabilistic quantifiers. 

Team NRU-HSE$_{4}$ uses an existing
quantification approach based on cost-sensitive learning, and adapted
it to the ordinal case. 

Team ISTI-CNR$_{6}$ instead used a novel
adaptation to quantification of a tree-based approach to ordinal
regression.

Teams LYS$_{7}$ and HSENN$_{9}$ also used an existing quantification
approach, but did not exploit the ordinal nature of the problem. 

The other teams mostly used approaches based on ``classify and count'' (see
Section \ref{sec:4Ddiscussion}), and viewed the problem as single-label
multi-class (instead of ordinal) classification; some of these teams
(notably, team Finki$_{2}$) obtained very good results, which testifies
to the quality of the (general-purpose) features and learning
algorithm they used.


\section{Conclusion and Future Work}
\label{sec:conclusion}

\noindent We described SemEval-2016 Task 4 ``Sentiment Analysis
in Twitter'', which included five subtasks including three that represent a
significant departure from previous editions. The three new subtasks
focused, individually or in combination, on two variants of the basic
``sentiment classification in Twitter'' task that had not been
previously explored within SemEval.  The first variant adopts a
five-point scale, which confers an \textit{ordinal} character to the
classification task.  The second variant focuses on the correct
estimation of the prevalence of each class of interest, a task which
has been called \textit{quantification} in the supervised learning
literature. In contrast, previous years' subtasks have focused on the
correct labeling of individual tweets.  As in previous years (2013--2015), the
2016 task was very popular and attracted a total of 43 teams.

A general trend that emerges from SemEval-2016 Task 4 is
that most teams who were ranked at the top in the various subtasks used
deep learning, including convolutional NNs, recurrent NNs, and 
(general-purpose or task-specific) word embeddings. In many cases, the
use of these techniques allowed the teams using them to obtain good
scores even without tuning their system to the specifics of the
subtask at hand, e.g., even without exploiting the ordinal nature of
the subtask -- for Subtasks C and E -- or the quantification-related
nature of the subtask -- for Subtasks D and E. Conversely, several
teams that have indeed tuned their system to the specifics of the
subtask at hand, but have not used deep learning techniques, have
performed less satisfactorily. This is a further confirmation of the
power of deep learning techniques for tweet sentiment analysis.

Concerning Subtasks D and E, if quantification-based subtasks are
proposed again, we think it might be a good idea to generate, for each
test topic $t_i$, multiple ``artificial'' test topics $t_i^1, t_i^2,
...$, where class prevalences are altered with respect to the ones of
$t_i$ by means of selectively removing from $t_i$ tweets belonging to
a certain class. In this way, the evaluation can take into
consideration (\emph{i}) class prevalences in the test set and
(\emph{ii}) levels of distribution drift (i.e.,~of the divergence of
the test distribution from the training distribution) that are not
present in the ``naturally occurring'' data. 

By varying the amount of
removed tweets at will, one may obtain \textit{many} test topics, thus
augmenting the magnitude of the experimentation at will while at the
same time keeping constant the amount of manual annotation needed.





In terms of possible follow-ups of this task, it might be interesting
to have a subtask whose goal is to distinguish tweets that are
\textsc{Neutral} about the topic (i.e., do not express any opinion
about the topic) from tweets that express a \textsc{Fair} opinion
(i.e., lukewarm, intermediate between \textsc{Positive} and
\textsc{Negative}) about the topic.

Another possibility is to have a multi-lingual tweet sentiment
classification subtask, where training examples are provided for the
same topic for two languages (e.g., English and Arabic), and where
participants can improve their performance on one language by
leveraging the training examples for the other language via transfer
learning. Alternatively, it might be interesting to include a
cross-lingual tweet sentiment classification subtask, where training
examples are provided for a given language (e.g., English) but not for
the other (e.g., Arabic); the second language could be also a surprise
language, which could be announced at the last moment.


\begin{table*}[htb]
  \begin{center}
    \resizebox{\textwidth}{!} {
    \begin{tabular}{|ccccc|l|l|l|l|}
      \hline
      \multicolumn{5}{|c|}{\textbf{Subtasks}} & \textbf{Team ID} & \textbf{Affiliation} & \textbf{Nation} & \textbf{Paper} \\ \hline 
      \hline \rule[0ex]{-1mm}{2ex} 
      A & & & & & Aicyber & Aicyber.com & Singapore; China & \cite{SemEval:2016:task4:Aicyber} \\ \hline
      A & & & & & aueb.twitter.sentiment & Department of Informatics, Athens University of Economics and Business & Greece &\cite{SemEval:2016:task4:aueb.twitter.sentiment} \\ \hline 
      \multirow{2}{*}{A} & & & & & \multirow{2}{*}{CICBUAPnlp} & Instituto Politècnico Nacional & \multirow{2}{*}{Mexico} & \multirow{2}{*}{\cite{SemEval:2016:task4:CICBUAPnlp}} \\ 
      & & & & & & Benemèrita Universidad Autonoma de Puebla & & \\ \hline 
      A & B & & & & CUFE & Cairo University & Egypt & \cite{SemEval:2016:task4:CUFE} \\ \hline 
      A & & & & & DIEGOLab16 & Arizona State University & USA & \cite{SemEval:2016:task4:DIEGOLab16} \\ \hline 
      A & & & & & DSIC-ELIRF & Universitat Politècnica de València & Spain & \cite{SemEval:2016:task4:DSIC-ELIRF} \\ \hline 
      A & B & C & D & E & ECNU & East China Normal University & China & \cite{SemEval:2016:task4:ECNU} \\ \hline 
      A & & & & & ELiRF & Universitat Politècnica de València & Spain &  \\ \hline 
      & B & C & D & E & Finki & 
      Saints Cyril and Methodius University, Skopje & Macedonia & \cite{SemEval:2016:task4:Finki} \\ \hline 
      A & B & & D & & GTI & AtlantTIC Centre, University of Vigo & Spain & \cite{SemEval:2016:task4:GTI} \\ \hline 
      & & & D & E & HSENN & National Research University Higher School of Economics & Russia &  \\ \hline 
      \multirow{2}{*}{A} & & & & & \multirow{2}{*}{I2RNTU} & Institute for Infocomm Research, A*STAR & \multirow{2}{*}{Singapore} & \multirow{2}{*}{\cite{SemEval:2016:task4:I2RNTU}} \\ 
      & & & & & & School of Computer Engineering, Nanyang Technological University & & \\ \hline 
      A & & & & & IIP & Infosys Limited & India & \cite{SemEval:2016:task4:IIP} \\ \hline 
      \multirow{2}{*}{A} & & & & & \multirow{2}{*}{INESC-ID} & INESC-ID, Lisboa & \multirow{2}{*}{Portugal} & \multirow{2}{*}{\cite{SemEval:2016:task4:INESC-ID}} \\ 
      & & & & & & Instituto Superior Técnico, Universidade de Lisboa & & \\ \hline 
      \multirow{2}{*}{A} & \multirow{2}{*}{B} & \multirow{2}{*}{C} & \multirow{2}{*}{D} & \multirow{2}{*}{E} & \multirow{2}{*}{INSIGHT-1} & INSIGHT Research Centre, National University of Ireland, Galway & \multirow{2}{*}{Ireland} & \multirow{2}{*}{\cite{SemEval:2016:task4:INSIGHT-1}} \\ 
      & & & & & & AYLIEN Inc. & & \\ \hline 
      \multirow{2}{*}{A} & \multirow{2}{*}{B} & \multirow{2}{*}{C} & \multirow{2}{*}{D} & \multirow{2}{*}{E} & \multirow{2}{*}{LYS} & Universidade da Coruña & \multirow{2}{*}{Spain} & \multirow{2}{*}{\cite{SemEval:2016:task4:LYS}} \\ 
      & & & & & & Universidade de Vigo & & \\ \hline 
      A & & & & & MDSENT & University of Maryland Baltimore County & USA & \cite{SemEval:2016:task4:MDSENT} \\ \hline 
      A & & & & & mib & Istituto di Informatica e Telematica, Consiglio Nazionale delle Ricerche & Italy & \cite{SemEval:2016:task4:mib} \\ \hline 
      A & & & & & Minions & University of Iasi & Romania & \cite{SemEval:2016:task4:Minions} \\ \hline 
      A & B & C & D & E & ISTI-CNR & Istituto di Scienza e Tecnologie dell'Informazione, Consiglio Nazionale delle Ricerche & Italy & \cite{SemEval:2016:task4:ISTI-CNR} \\ \hline 
      & & & D & E & NRU-HSE & National Research University Higher School of Economics & Russia & \cite{SemEval:2016:task4:NRU-HSE} \\ \hline 
      A & & & & & NTNUSentEval & Norwegian University of Science and Technology & Norway & \cite{SemEval:2016:task4:NTNUSentEval} \\ \hline 
      A & B & & & & OPAL & European Commission Joint Research Centre & Italy & \cite{SemEval:2016:task4:OPAL} \\ \hline 
      & B & C & D & E & pkudblab & Peking University & China &  \\ \hline 
      & \multirow{2}{*}{B} & \multirow{2}{*}{C} & \multirow{2}{*}{D} & \multirow{2}{*}{E} & \multirow{2}{*}{PotTS} & University of Potsdam & \multirow{2}{*}{Germany} & \multirow{2}{*}{\cite{SemEval:2016:task4:PotTS}} \\ 
      & & & & & & Retresco GmbH & & \\ \hline 
      A & B & C & & & PUT & Poznan University of Technology & Poland & \cite{SemEval:2016:task4:PUT} \\ \hline 
      & & & D & E & QCRI (**) & Qatar Computing Research Institute & Qatar & \cite{DaSanMartino:2016ty} \\ \hline 
      A & & & & & SENSEI-LIF & Aix-Marseille University - CNRS - LIF & France & \cite{SemEval:2016:task4:SENSEI-LIF} \\ \hline 
      A & B & C & & & SentimentalITists & University of Iasi & Romania & \cite{SemEval:2016:task4:SentimentalITists} \\ \hline 
      A & & & & & SentiSys & Aix-Marseille University & France & \cite{SemEval:2016:task4:SentiSys} \\ \hline 
      \multirow{3}{*}{A} & & & & & \multirow{3}{*}{SteM} & Sabanci University & \multirow{2}{*}{Turkey} & \multirow{3}{*}{\cite{SemEval:2016:task4:SteM}} \\ 
      & & & & & & Marmara University & &  \\
      & & & & & & Otto-von-Guericke University Magdeburg & Germany &  \\ \hline 
      A & B & & D & & SwissCheese & ETH Zürich & Switzerland & \cite{SemEval:2016:task4:SwissCheese} \\ \hline 
      & B & & & & thecerealkiller & Amazon.in & India & \cite{SemEval:2016:task4:thecerealkiller} \\ \hline 
      A & & & & & THUIR & Tsinghua University & China &  \\ \hline 
      \multirow{4}{*}{A} & \multirow{4}{*}{B} & & & & \multirow{4}{*}{Tweester} & School of ECE, National Technical University of Athens & \multirow{5}{*}{Greece} & \multirow{5}{*}{\cite{SemEval:2016:task4:Tweester}} \\ 
      & & & & & & School of ECE, Technical University of Crete & & \\ 
      & & & & & & Department of Informatics, University of Athens & &  \\ 
      & & & & & & Signal Analysis and Interpretation Laboratory (SAIL) & &  \\ 
      & & & & & & Institute for Language \& Speech Processing - ILSP & &  \\ \hline 
      A & B & C & D & & TwiSE & University of Grenoble-Alpes & France & \cite{SemEval:2016:task4:TwiSE} \\ \hline 
      & & & D & & UDLAP & Universidad de las Américas Puebla (UDLAP) & Mexico & \cite{SemEval:2016:task4:UDLAP} \\ \hline 
      A & B & & & & UNIMELB & University of Melbourne & Australia & \cite{SemEval:2016:task4:UNIMELB} \\ \hline 
      A & & & & & UniPI & Università di Pisa & Italy & \cite{SemEval:2016:task4:UniPI} \\ \hline 
      A & & & & & UofL & University of Louisville & USA & \cite{SemEval:2016:task4:UofL} \\ \hline 
      A & B & & & & VCU-TSA & Virginia Commonwealth University & USA & \cite{SemEval:2016:task4:VCU-TSA} \\ \hline 
      A & B & & & & WR & WR & Hong Kong 
      &  \\ \hline 
      & & \multirow{2}{*}{C} & & & \multirow{2}{*}{YZU-NLP} & Yuan Ze University, Taoyuan & Taiwan & \multirow{2}{*}{\cite{SemEval:2016:task4:YZU-NLP}} \\
      & & & & & & Yunnan University, Kunming & China &  \\ \hline 
      \hline\rule[0ex]{0mm}{2ex} 
      \textbf{34} & \textbf{19} & \textbf{11} & \textbf{14} & \textbf{10} & \textbf{Total} & & & \\ \hline
    \end{tabular}
    }
  \end{center}
  \caption{Participating teams (Column 2), their affiliation (Column 3) and nationality (Column 4), the subtasks they have participated in (Column 1), and the paper they have contributed (Column 5). Teams whose ``Affiliation'' column is typeset on more that one row include researchers with different affiliations. Teams marked with a (**) include some of the SemEval 2016 Task 4 organizers. An empty entry for the ``Paper'' column indicates that the team have not contributed a system description paper. \label{tab:participants}}
\end{table*}




\bibliographystyle{naaclhlt2016}
\bibliography{Preslav,Alan,Sara,Fabrizio,Veselin,SemEvalBiblio,systems}

\end{document}